# Dehumanizing Voice Technology: Phonetic & Experiential Consequences of Restricted Human-Machine Interaction

**Christian Hildebrand,[1] Donna Hoffman,[2] Tom Novak[2]**

University of St. Gallen,[1] George Washington University[2]

christian.hildebrand@unisg.ch,[1] dlhoffman@gwu.edu,[2] novak@gwu.edu[2]

### Abstract

The use of natural language and voice-based interfaces gradually transforms how consumers search, shop, and express their preferences. The current work explores how changes in the syntactical structure of the interaction with conversational interfaces (command vs. request based expression modalities) negatively affects consumers' subjective task enjoyment and systematically alters objective vocal features in the human voice. We show that requests (vs. commands) lead to an increase in phonetic convergence and lower phonetic latency, and ultimately a more natural task experience for consumers. To the best of our knowledge, this is the first work documenting that altering the input modality of how consumers interact with smart objects systematically affects consumers' IoT experience. We provide evidence that altering the required input to initiate a conversation with smart objects provokes systematic changes both in terms of consumers' subjective experience and objective phonetic changes in the human voice. The current research also makes a methodological contribution by highlighting the unexplored potential of feature extraction in human voice as a novel data format linking consumers' vocal features during speech formation and their subjective task experiences.

## Introduction

The use of natural language and voice-based interfaces gradually transforms how consumers interact with firms (Dale 2016; Hirschberg and Manning 2015). The use of voice-based interfaces as a new interaction paradigm between human consumers and intelligent bots (such as Amazon Alexa, Google Home, or Siri) has been declared as the "next operating system in commerce" (Feldman, Goldenberg, and Netzer 2010; Suri, Elia, and van Hillegersberg 2017). With 100 million smart speakers installed in people's home worldwide in 2018 and a soaring market of voice assistant technologies that is expected to reach $31.82 billion by 2025, voice-based interfaces are transforming how humans search, shop, and express their preferences.

## The Unexplored Role of Voice-Mediated Response Modalities

Research on the impact of voice-based interaction modalities on consumers is both scarce and predominantly concerned with design, security, or general technology-acceptance issues rather than the consequences for consumer behavior. Specifically, the majority of prior work on voice-based or "conversational" interfaces primarily examined either factors related to optimizing the design features of interfaces (Ghosh and Pherwani 2015), factors related to security issues of voice-controlled interfaces (Diao et al. 2014), or general user acceptance (Portet et al. 2013).

The current work takes a different route and explores whether the input or task initiation modality (i.e., how consumers are required to talk to a conversational interface) can systematically alter consumers' subjective task experience and their underlying attributions toward the interface. Building on recent conceptual foundations of master-servant relationships expressed during IoT interaction events (Hoffman and Novak 2018; Novak and Hoffman 2018), the current work examines how restricted task initiation modalities during human-object interaction can evoke more negative task experiences, more negative attributions toward a voice-based assistant, and systematic changes in the human voice.

## Experimental Paradigm & Procedure

To explore whether and how variation in task initiation modalities affects consumers' task experience, we recruited a total of 100 participants for a laboratory study in exchange for monetary compensation ($M_{Age}$=24.27, $SD_{Age}$=6.28, 51% females). At the outset of the study, we assessed participants' baseline vocal features using an established reading task from prior work in bioacoustics (Kempster et al. 2009; example sentence: "The blue spot is on the key again."). Next, participants were randomly assigned to either a restricted (commands) or a non-restricted (requests) task initiation modality condition. In both conditions, participants were provided with a set of eleven commands which required them to engage in a turn-taking exercise with Amazon Alexa. However, the two conditions differed systematically in the level of restrictedness: In the restricted command interaction modality condition participants received eleven commands that required a syntactically shortened form of interaction (example command: "Alexa, length of marathon."). In the non-restricted request condition, participants received the same substantive commands but all requests were expressed as in a non-restricted, more natural conversation (example command: "Alexa, can you tell me the length of a marathon?"). All tasks were carefully selected, leading to identical responses of the voice assistant independent of the task initiation modality (i.e., providing identical answers in all cases except for minor variations in one question related to telling a joke). Immediately after the voice-assistant task was completed, we assessed the perceived naturalness of the interaction (7-point Likert scale, example item: "The interaction with this interface felt …. natural / robotic(r) / lacking depth (r)", $\alpha_{natural}$=.79), participants' overall task enjoyment ("This task was a lot of fun"), attributions of competence ("This systems is … an expert / competent / proficient", $\alpha_{comp}$=.81) and warmth ("This systems is … … compassionate / sympathetic / warm", $\alpha_{warmth}$=.85), and a final set of demographic questions. The audio data of both the baseline reading task and the actual object-interaction task was recorded using an external BlueYeti microphone with a predefined sampling rate of 44100 HZ. Processing of all audio data and extraction of vocal features at the participant level was done using the seewave and tuneR packages in R (Sueur, Aubin, and Simonis 2008).

## Results

As demonstrated in Figure 1, our findings revealed that a restricted task initiation led to significantly less natural perceptions of the voice-assistant interaction ($M_{Restrict}$=3.70, $M_{NonRestrict}$=4.43; t(98)=2.654, $p<.01$). Furthermore, participants in the restricted task initiation condition also perceived the task as significantly less enjoying compared to the non-restricted initiation condition ($M_{Restrict}$=4.18, $M_{NonRestrict}$=4.96; t(98)=1.675, $p=.09$). A single mediation model with bootstrapped estimates confirmed that the effect on task enjoyment was offset after controlling for the effect of perceived naturalness on task enjoyment (5000 bootstrap resamples; $CI_{95\%}$=[.07;.94]), indicating full mediation. Greater naturalness was also a key factor predicting an increase in attributions of both competence ($\beta_{Natural}$=.36, t(98)=3.83, $p<.001$) and warmth ($\beta_{Natural}$=.74, t(98)=8.60, $p<.001$).

If greater naturalness alone is sufficient to explain the increase in consumers' subjective task experience, then including the fitted values of attributions toward the voice assistant should provide no significant improvement. However, a series of Davidson MacKinnon tests for non-nested models revealed that including the fitted values of the attribution dimensions (competence and warmth) led to a significant improvement of model fit ($\theta_{Natural+Fitted(CompWarmthModel)}$ = .92, t = 4.413, $p<.001$) whereas including the fitted values of the naturalness only model did not further improve the fit of the model ($\theta_{CompWarmth+Fitted(NaturaModel)}$ = .17, t = 0.575, $p>.56$), indicating that the attributions toward the system (evoked through perceptions of a more natural task experience) led to the revealed increase in task enjoyment.

Furthermore, the manipulation of task initiation also significantly altered participants' vocal expressions. Specifically, and as shown in Figure 2, we find that a restricted compared to a non-restricted task initiation led to an increase in sound pressure levels both relative to the baseline measure of the reading task (t(99)=10.917, $p<.001$) as well as the between-subjects comparison of vocal features during the object-interaction task ($M_{Restrict}$=8.9052, $M_{NonRestrict}$=8.9044; t(98)=3.242, $p<.01$). Similarly, restricted relative to non-restricted task initiation also significantly increased participants' vocal entropy, again, both relative to the baseline measure of the reading task (t(99)=13.819, $p<.001$) as well as during the object-interaction task ($M_{Restrict}$=.69, $M_{NonRestrict}$=.66; t(98)=4.167, $p<.001$). These findings suggest that a restricted task initiation causes individuals to engage in a higher pitch and a less fluent (or more monotone) vocal expression during speech formation. This change in individuals' vocal features was also reflected in a significant, negative correlation between perceptions of naturalness and vocal entropy (r(98)=-.64, $p<.001$), confirming that the less fluent vocal expression was also associated with lower task enjoyment. Sound pressure levels were directionally consistent but non-significant (i.e., negative correlation with r(98)=-.07, p=.50).

**Figure 1. Negative Consequences of Command-Based Task Initation**

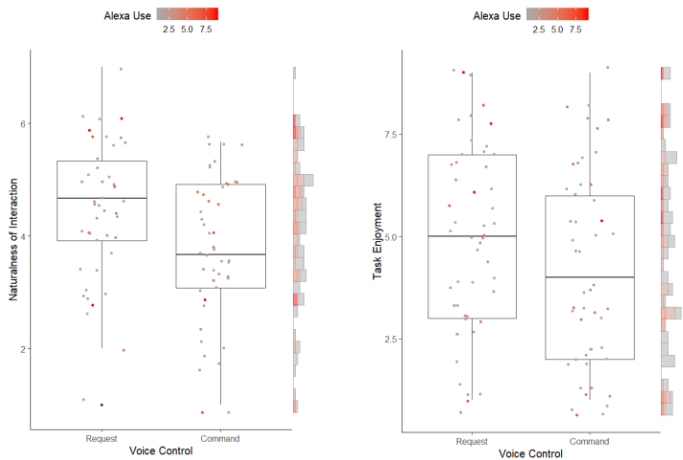

**Figure 2. Phonetic Shifts Relative to Baseline**

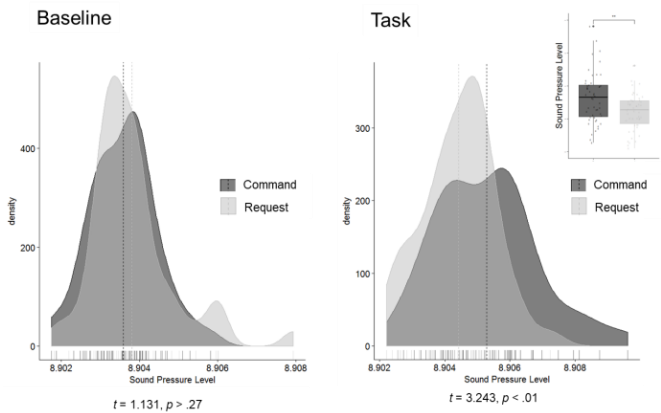

## Conclusion

To the best of our knowledge, this is the first work documenting that altering the input modality of how consumers interact with smart objects systematically affects consumers' IoT experience. We provide evidence that altering the required input to initiate a conversation with smart objects provokes systematic changes of both consumers' subjective experience and also objective phonetic changes in the human voice. The current research also makes a methodological contribution by highlighting the unexplored potential of feature extraction in the human voice as a novel data format linking consumers' vocal features during speech formation and their subjective task experiences during human-machine interactions.

## References

Dale, Robert (2016), "The return of the chatbots," *Natural Language Engineering*, 22 (5), 811–17.

Diao, Wenrui, Xiangyu Liu, Zhe Zhou, and Kehuan Zhang (2014), "Your Voice Assistant is Mine," in *Proceedings of the 4th ACM Workshop on Security and Privacy in Smartphones & Mobile Devices - SPSM '14*, 63–74.

Feldman, Ronen, Jacob Goldenberg, and Oded Netzer (2010), "Mine your own business: Market structure surveillance through text mining," *Marketing Science Institute, Special Report*, 10 (3), 10–202.

Ghosh, S and J Pherwani (2015), "Designing of a Natural Voice Assistants for Mobile Through User Centered Design Approach," in *Human-Computer Interaction - Design and Evaluation*, 320–31.

Hirschberg, Julia and Christopher D Manning (2015), "Advances in natural language processing," *Science*, 349 (6245), 261–66.

Hoffman, Donna L. and Thomas P. Novak (2018), "Consumer and object experience in the internet of things: An assemblage theory approach," *Journal of Consumer Research*, 44 (6), 1178–1204.

Kempster, Gail B., Bruce R. Gerratt, Katherine Verdolini Abbott, Julie Barkmeier-Kraemer, and Robert E. Hillman (2009), "Consensus auditory-perceptual evaluation of voice: Development of a standardized clinical protocol," *American Journal of Speech-Language Pathology*, 18 (2), 124–32.

Novak, Thomas P. and Donna L. Hoffman (2018), "Relationship journeys in the internet of things: a new framework for understanding interactions between consumers and smart objects," *Journal of the Academy of Marketing Science*.

Portet, François, Michel Vacher, Caroline Golanski, Camille Roux, and Brigitte Meillon (2013), "Design and evaluation of a smart home voice interface for the elderly: Acceptability and objection aspects," *Personal and Ubiquitous Computing*, 17 (1), 127–44.

Serban, Iulian V., Chinnadhurai Sankar, Mathieu Germain, Saizheng Zhang, Zhouhan Lin, Sandeep Subramanian, Taesup Kim, Michael Pieper, Sarath Chandar, Nan Rosemary Ke, Sai Rajeshwar, Alexandre de Brebisson, Jose M. R. Sotelo, Dendi Suhubdy, Vincent Michalski, Alexandre Nguyen, Joelle Pineau, and Yoshua Bengio (2017), "A Deep Reinforcement Learning Chatbot," *ArXiv preprint*, 1–34.

Sueur, Jerome, Thierry Aubin, and Caroline Simonis (2008), "Equipment review: Seewave, a free modular tool for sound analysis and synthesis," *Bioacoustics*, 18 (2), 213–26.

Suri, Vipin K., Marianne Elia, and Jos van Hillegersberg (2017), "Software bots -The next frontier for shared services and functional excellence," in *Lecture Notes in Business Information Processing*, 81–94.